\documentclass{article}
\usepackage{spconf,amsmath,graphicx}

\usepackage{amsfonts}
\usepackage{graphicx}
\usepackage{textcomp}
\usepackage{xcolor}

\usepackage{bm}
\usepackage[noend]{algpseudocode}
\usepackage{algorithm}

\title{
	NAMF: A Non-local Adaptive Mean Filter for Salt-and-Pepper Noise Removal
}

\name{Houwang Zhang$^{1,2}$ \qquad Yuan Zhu$^{1,2, \LARGE{*}}$ \qquad Hanying Zheng$^{1,2}$}

\address{$^{1}$ China University of Geosciences, School of Automation, Wuhan, China \\
$^{2}$ Hubei Key Laboratory of Advanced Control and Intelligent Automation for Complex Systems, Wuhan, China \\
$^{\LARGE{*}}$ Corresponding author: Yuan Zhu (e-mail: zhuyuan@cug.edu.cn)
}

\begin{document}

\maketitle

\begin{abstract}
In this paper, a novel algorithm called a non-local adaptive mean filter (NAMF) for removing salt-and-pepper (SAP) noise from corrupted images is presented. We employ an efficient window detector with adaptive size to detect the noise, the noisy pixel will be replaced by the combination of its neighboring pixels, and finally we use a SAP noise based non-local mean filter to reconstruct the intensity values of noisy pixels. Extensive experimental results demonstrate that NAMF can obtain better performance in terms of quality for restoring images at all levels of SAP noise.
\end{abstract}

\section{Introduction}
\label{sec:intro}

Digital images are often corrupted by noises in the process of image acquisition and transmission \cite{Chandra2011Removal,Luo2006Efficient}. While the existence of noises will make tasks of image processing and computer vision become seriously ill-posed problems \cite{ma2}. As a pre-processing step in image processing, image denoising can protect edges, textures and other details \cite{ma3}. Hence it is taken as one of the most important tasks in image processing \cite{Xu2017A,goyal2020image}. Salt-and-pepper (SAP) noise commonly exists in natural images, and the pixels contaminated by SAP noise take the maximum or minimum value and can be represented as black or white points \cite{singh2018adaptive,karthik2020removal}. 

To remove SAP noise, lots of computational methods have been proposed. Among them, median filter (MF) and adaptive median filter (AMF) \cite{Hwang1995Adaptive} are the two most popular methods in the early stage. MF can restore image details well under low noise intensity, but it performs poorly when noise intensity is high \cite{Chan2005Salt}. AMF adopts the measure of window with adaptive size, which makes it perform well in high noise intensity \cite{deng2016new}.

In recent years, studies on denoising of SAP noise are mainly based on MF and AMF. Some researches also use deep learning methods. However, these deep learning methods depend on data \cite{he2018novel}. Based on AMF, Noise adaptive fuzzy switching median filter (NAFSMF) recognizes SAP noise by analyzing the histogram of noisy images and takes a fuzzy method to denoise \cite{Toh2010Noise}. Adaptive weighted mean filter (AWMF) uses two successive windows to detect noisy pixels and processes them with a weighted mean filter \cite{Zhang2014A}. The method proposed in \cite{Yi2016An} is based on NAFSMF and AWMF and uses a new adaptive fuzzy switching weighted mean filter to remove SAP noise. In \cite{tbtkelektrik400946}, researchers proposed a method, based on pixel density filter (BPDF), to remove SAP noise through searching the repeated numbers of the pixels, and achieves a good performance under low SAP noise intensity. Different applied median filter (DAMF) is proposed for removing SAP noise at all densities \cite{ERKAN2018789}.

\begin{figure}[htbp]
\centerline{\includegraphics[width=1\columnwidth]{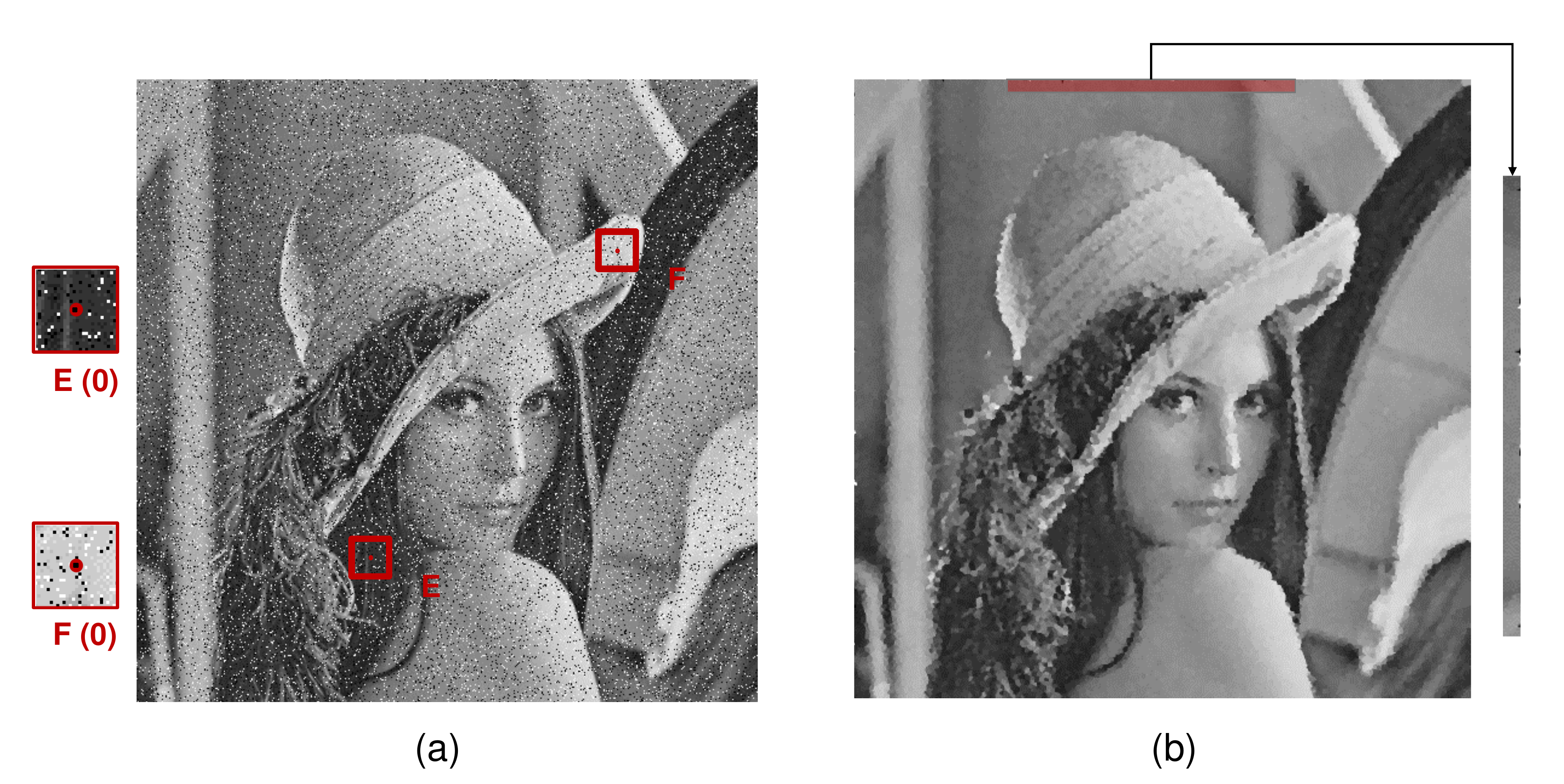}}
\caption{(a) "lena" corrupted by SAP noise ratio of 10\%, and the gray values of noisy candidates in $E$ and $F$ are both $0$. (b) Result of the method proposed  in \cite{Yi2016An} for "Lena" corrupted by SAP noise ratio of 90\%.}
\label{fig1}
\end{figure}

The restoration of SAP noise is just to use the rest information (uncontaminated pixels) to repair the absent information (contaminated pixels). As Fig. \ref{fig1} (b) shows, the limitation of existing state-of-the-art methods is that under high SAP noise level the boundary of the restored image is jagged and the details are blurred. When SAP noise intensity is too high, the image loses too much information. The consequence is that less information is available to be used for restoring the image.

To solve the above problem, in this paper, we propose a non-local adaptive mean filter (NAMF) to remove SAP noise efficiently. NAMF can be divided into two stages: (1) SAP noise detection; (2) SAP noise elimination. Firstly, in the stage of SAP noise detection, we find out pixels whose gray value are equal to the global maximum or minimum gray value of the image. We then take them as noisy candidates and use a window with adaptive size to confirm them. For the possible noisy pixel, we calculate the proportion of the pixels with the same value of candidate in the window, and then filter it by a threshold. If it is smaller than the threshold, the candidate is regarded as a noisy pixel, otherwise it is noiseless and will not be processed. Secondly, in the stage of SAP noise elimination, the noisy pixel will be replaced by the mean of its neighboring pixels. Then we use a SAP noise based non-local mean filter to further restore it. The main contributions of this paper can be concluded as follows:

\begin{itemize}
\item A proportion based method is proposed to distinguish the noise pixels and texture pixels in the SAP noise detection, which can avoid noiseless texture pixels being processed as noisy pixels.

\item An improved non-local mean method based on the characteristics of SAP noise is raised to further restore noisy images, which can further enhance the quality of restored images.
\end{itemize}

Experimental results on 16 typical images and 40 test images in the TEST IMAGES Database \cite{testingimge} demonstrate that NAMF outperforms the existing state-of-the-art methods under both high SAP noise level and low SAP noise level \footnote{All of the work (data and codes for our proposed method) is published on https://github.com/ProfHubert/NAMF.}.

\section{Related Works}

In this section we will firstly review traditional MF based methods used for SAP noise removal and then we introduce the classical Non-local means method.

\subsection{MF based Methods for SAP Noise Elimination}

MF is the most commonly used algorithm to remove pulse interference and SAP noise. The main idea is that all pixels are replaced by neighborhood median pixels whether pixels are noiseless or not. Detailedly, let $\bm{X}$ be an original 8-bit gray level image with size of $M \times N$, and $x_{i,j}$ be the gray value of the pixel at location $(i, j)$. Then, for MF with filter size of $3 \times 3$, the new gray value $\tilde{x}_{i,j}$ of pixel at location $(i, j)$ can be calculated by Eq. (\ref{eq0}).

\begin{equation}
\begin{split}
\label{eq0}
\tilde{x}_{i,j}=median( x_{i-1, j-1}, x_{i-1, j}, x_{i-1, j+1}, x_{i, j-1}, x_{i,j}, \\ x_{i, j+1}, x_{i+1, j-1}, x_{i+1, j}, x_{i+1, j+1})
\end{split}
\end{equation}

Based on the idea of MF, many MF methods have been proposed. However, when the distance between the noisy pixel and its neighborhood is large, the difference of the median value pixel and the original pixel that is noiseless may be large, and that would cause the denoised image to be degraded \cite{Yi2016An}. In \cite{Zhang2014A}, the median value of neighborhoods is combined with different weights. For different neighborhoods, weights are computed based on the distance from the center noisy pixel. Besides, when noise density is too high, for some noisy pixels, there are no noiseless neighborhoods for restoring. In \cite{Yi2016An}, to tackle the problem, the researchers provide a method by using neighborhoods in the restored image to calculate the median value.

\subsection{Non-local Mean (NLM) Method}

The NLM method combined the idea of the yaroslavsky filter \cite{yaroslavsky2012digital} and bilateral filter \cite{tomasi1998bilateral}. For the noisy pixel, NLM use the intensity patch feature instead of single pixel feature to compute the denoised mean value \cite{Buades2005A}.

%Given a pixel $x \in X$, $N(x)$ denotes an image block centered at x with size $p \times p$. Then, NLM calculates a weighted average of all the pixels in the searching window $I$ by 
%
%\begin{equation}
%\tilde{x}=
%\frac{1}{C_i}\sum_{j\in B}x(i)u(i,j)
%\end{equation}

Given a pixel $x \in \bm{I}$, let $N(x)$ denote an image block centered at $x$. Then, NLM calculates a weighted average of all the pixels in the searching window $B$ by Eqs. (\ref{n1}) - (\ref{n3}).

\begin{equation}
\label{n1}
\tilde{x}=
\frac{\sum_{y \in B} y * u(x, y)}{\sum_{y \in B} u(x, y)}
\end{equation} 

\begin{equation}
\label{n2}
\begin{split}
u(x,y)=
e^{-\frac{d(x,y)}{h^{2}}}
\end{split}
\end{equation}

\begin{equation}
\label{n3}
d(x,y)=\left \| N(x) - N(y) \right \|^{2}_{2,a}
\end{equation}
where $\tilde{x}$ is the restored gray value of $x$. $u(x, y)$ denotes the similarity between pixels $x$ and $y$, which depends on the weighted Euclidean distance between two image blocks $N(x)$ and $N(y)$. $h$ is the smoothing parameter, $a$ is the standard deviation of the Gaussian kernel.

\section{The Proposed Non-loacl Adaptive Mean (NAMF) Filter}

In this section we will present the details of our proposed non-local adaptive mean (NAMF) filter. As the most traditional SAP noise removal filters, NAMF can be divided into two stages: (1) SAP noise detection; (2) SAP noise elimination. Details are presented as follows.

\subsection{SAP Noise Detection}

In accordance with the mathematical notation as the previous section, $\bm{X} = (x_{i,j}) \in \mathbb{R}^{M \times N}$ and $\bm{Y} = (y_{i,j}) \in \mathbb{R}^{M \times N}$ represent the original 8-bit gray-level image and noisy image corrupted by SAP noise, respectively. $x_{i,j}$ and $y_{i,j}$ represent the gray value of the pixel at location $(i, j)$ of $\bm{X}$ and $\bm{Y}$, respectively, where $(i,j) \in \Lambda  \equiv \left \{1, \cdots ,M \right \} \times \left \{1, \cdots , N \right \}$.

In a corrupted image, the value of a "salt" pixel equals to the maximum gray value 255, and the value of a "pepper" pixel equals to the minimum value 0. Thus, $y_{i,j}$ is defined by Eq. (\ref{eq1}).

\begin{equation}
\label{eq1}
y_{i,j}=
\begin{cases}
255 \, \, or \, \,  0, & \text{with probability $\alpha$}\\
x_{i,j}, & \text{with probability $1-\alpha$}
\end{cases}
\end{equation} 
where $\alpha$ is the density of SAP noise of $\bm{Y}$. And in the process of image denoising, we use $S_{i,j}(w)$ to represent a $(2w + 1) \times (2w + 1)$ window centered at $(i,j)$ with the radius $w$.

Considering the characteristics of SAP noise, pixel $y_{i,j}$ corrupted by SAP noise is $0$ or $255$. That is to say, noisy pixel candidate $y_{i,j}$ only has 2 possible values: $y_{min} = 0$ and $y_{max} = 255$. Following is a prior decision condition in noise detection.

\begin{equation}
\label{eq2}
o(i, j)=
\begin{cases}
1,& \text{$y_{i,j} = 0 \,\, or \,\, 255$}\\
0,& \text{$otherwise$}
\end{cases}
\end{equation}
where $\bm{O} = (o_{i,j}) \in \mathbb{R}^{M \times N}$ is an indicator matrix with binary value. $o(i,j)=1$ means that pixel $y_{i,j}$ is the noisy pixel candidate, while $o(i,j)=0$ that means pixel $y_{i,j}$ is noiseless. For a natural image, the pixels with high or low value are also possible to be the texture of the image. For example, as shown in Fig. \ref{fig1} (a), $E$ and $F$ are both candidates with gray value of $0$ (black). However, $F$ has a larger possibility to be a noisy pixel, and $E$ has a larger possibility to be a texture pixel. Therefore, it is very necessary to further confirm these pixels with maximum or minimum value.

In fact, in a natural image pixels are supposed to change smoothly, and adjacent pixels tend to have similar values. The value of pixels in an image with SAP noise may change suddenly. Motivated by these facts, we propose an adaptive SAP noise detection method by considering the neighborhood to identify noise candidates as noisy pixels or not. Its procedures can be divided into two steps.

\begin{itemize}
	\item $\bm{Step \ 1:}$ For the pixel $y_{i,j}$ with $o_{i, j} = 1$, we calculate $S^{sum}_{i,j}(w)$ within an adaptive searching window $S_{i, j}(w)$. The radius $w$ of window $S_{i, j}(w)$ is initialed to 1. If the condition is met ($w == w_{max}$ or $S^{sum}_{i,j}(w) > 0$), the $w$ is just we need. If not, then $w + 1$ and continue to compute, where $S^{sum}_{i,j}(w)$ is the number of pixels within $S_{i,j}(w)$ which are not equal to $y_{min}$ and $y_{max}$, $w_{max}$ is the maximum size of window. If $S^{sum}_{i,j}(w) > 0$, we take the pixel $y_{i,j}$ as a noisy pixel, the detection for $y_{i,j}$ ends. Otherwise we go to $\bm{Step \ 2}$.
	
	\item $\bm{Step \ 2:}$  If  $S^{sum}_{i,j}(w) = 0$, $y_{i,j}$ maybe considered as a texture pixel. Then we calculate the proportion $\rho = \frac{S^{num}_{i,j}(w)}{(2w + 1) \times (2w + 1))}$, where $S^{num}_{i,j}(w)$ is the number of pixels owning same value as candidate pixel $y_{i, j}$ in the window $S_{i,j}(w)$. After that we set a threshold $T$ to identify the noisy pixel. If $\rho \leq T$, the candidate pixel $y_{i, j}$ is regards as a noisy pixel, else it is noiseless.
\end{itemize}

%In the detection of SAP noise, size of window $w$ is initialed to 1. If the condition is met ($w == w_{max}$ or $S^{sum}_{i,j}(w) > 0$), the $w$ is just we need. If not, then $w + 1$ and continue to compute, where $S^{sum}_{i,j}(w)$ is the number of pixels within $S_{i,j}(w)$ which are not equal to $y_{min}$ and $y_{max}$, $w_{max}$ is the maximum size of window. After the computation of $w$, if $S^{sum}_{i,j}(w) > 0$, we take the pixel in the window as a noisy pixel. If not, it may be a texture pixel, then we compare the proportion $P = \frac{S^{num}_{i,j}(w)}{(2w + 1) \times (2w + 1))}$ with the threshold $T$, where $S^{num}_{i,j}(w)$ is the number of pixels owning same value as candidate pixel $y_{i, j}$ in the window $S_{i,j}(w)$. If $P \leq T$, the candidate pixel $y_{i, j}$ is regarded as a noisy pixel, else it is noiseless.

For the pixel $y_{i, j}$, if it is finally detected as noisy, we mark it with the discriminant matrix $\bm{L} = (l_{i,j}) \in \mathbb{R}^{M \times N}$, and $l(i, j) = 1$, else $l(i, j) = 0$ and will not be processed.

\subsection{SAP Noise Elimination}

We restore noisy pixels in two steps. Let $\bm{Z}$ represent the initially restored image, $\bm{\hat{Z}}$ represent the final output image. Before processing, we initialize $\bm{Z}$ with $\bm{Z} = \bm{Y}$ .

\begin{figure}[htbp]
	\centerline{\includegraphics[width=1\columnwidth]{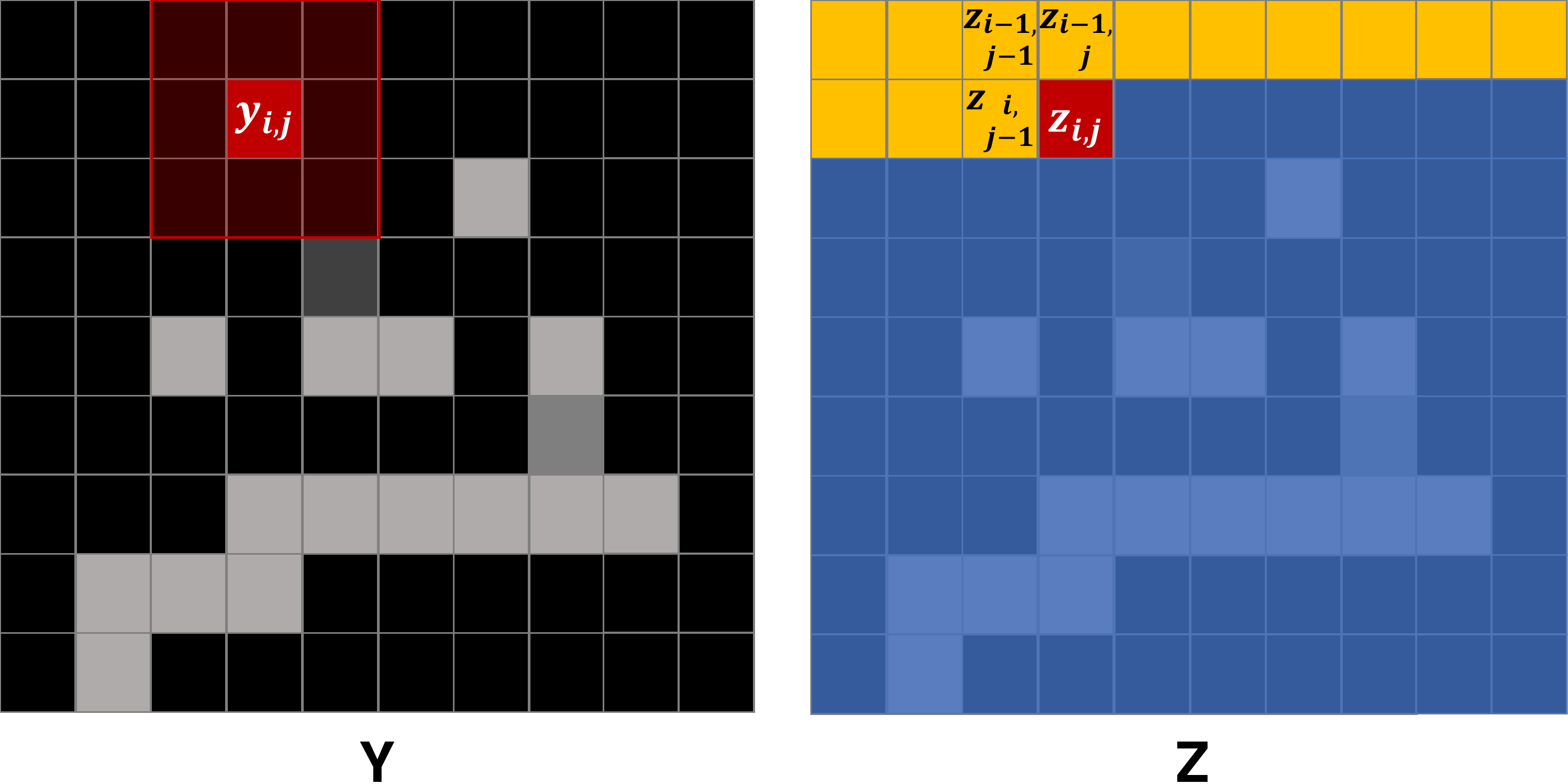}}
	%	\caption{Three neighboring pixels $z_{i-1,j-1}, z_ {i-1,j}$ and $z_{i,j-1}$ are selected in restored image $z$ for restoring pixel $y_{i, j}$ in original image $y$ when $S^{sum}_{i,j}(w) = 0$. $y$ is a noisy image, $z$ is a image being restored, the red pixel in $z$ is prepared to process, the yellow pixels are pixels which have been restored, other blue pixels in $z$ are pixels which remain to be restored.}
	
	\caption{$\bm{Y}$ is a noisy image, $\bm{Z}$ is an image being restored, the red pixel in $\bm{Z}$ is prepared to process, the yellow pixels are pixels which have been restored, other blue pixels in $\bm{Z}$ are pixels remained to be restored. When processing noisy pixel $y_{i,j}$, neighborhoods in the window are all noisy, i.e. $S^{sum}_{i,j}(w) = 0$. So here we select three neighboring pixels $z_{i-1,j-1}, z_ {i-1,j}$ and $z_{i,j-1}$ in restored image $\bm{Z}$ for restoring pixel $y_{i, j}$ in original image $\bm{Y}$.}
	
	\label{fig7}
\end{figure}

Firstly, when a pixel $y_{i,j}$ is detected as a noisy pixel, we use $S^{mean}_{i,j}(w)$ to restore it. The calculation of $S^{mean}_{i,j}(w)$ is based on $S^{sum}_{i,j}(w)$ as shown in Eq. (\ref{eq3}). When $S^{sum}_{i,j}(w) \neq 0$, $S^{mean}_{i,j}(w)$ is the mean of the noiseless pixels in $S_{i,j}(w)$, otherwise is the mean of three processed neighboring pixels of $y_{i,j}$ in $\bm{Z}$. Unlike the four neighbors adopted in \cite{Yi2016An}, the utilization of four neighboring pixels will lead to residual noisy pixels on the boundary, as shown in Fig. \ref{fig1}. (b) (even if the image boundary is expanded during process, some noisy pixels located in boundary can still not be restored). Hence, we select three processed neighboring pixels in $\bm{Z}$, i.e., $S^{mean}_{i,j}(w) = (z_{i-1,j-1} + z_ {i-1,j} + z_{i,j-1})/3$, and its illustration is shown in Fig. \ref{fig7}.

\begin{equation}
\label{eq3}
S^{mean}_{i,j}(w) \! = \!
\begin{cases} \!
\frac{\sum_{(e, f) \in S_{i,j}(w)} (1 \! - \! l(e,f))* y_{e,f}}{\sum_{(e, f) \in S_{i,j}(w)}  (1 - l(e, f))}, \! & \! \text{$S^{sum}_{i,j}(w) \! \neq \! 0$}\\
\frac{z_{i-1,j-1} + z_ {i-1,j} + z_{i,j-1}}{3}, \! & \! \text{$otherwise$}
\end{cases}
\end{equation}

Secondly, considering the problem that restoration will be hard under high noise intensity, we introduce the non-local mean (NLM) method \cite{Buades2005A,coupe2008bayesian} to further restore noisy pixels. NLM can restore noisy pixels by using all neighbors instead of part of neighbors which are detected as noiseless. In this way we can use NLM to further enhance the restoration effect. Considering that original NLM is designed for Gaussian noise or Speckle noise, which are quite different from SAP noise. We modify it based on the characteristics of SAP noise and make it applicable for our method.

In the noise detection stage, pixel is identified and marked by the discriminant matrix $\bm{L}$. $l(i, j)=1$ means the pixel $y_{i,j}$ is identified as a noisy pixel, otherwise noiseless. After the restoration of the noisy pixel $y_{i, j}$, we get $z_{i, j}$. The modified NLM method is used to further restore $z_{i,j}$. We use $\hat{z}_{i,j}$ to represent the value of the further restored $z_{i,j}$, and it can be calculated as follows:

\begin{equation}
\label{eq4}
\hat{z}_{i, j}=
\begin{cases}
\frac{1}{C}\sum_{z_{e,f}\in B(z_{i,j}, \ r)}z_{e,f} * u(z_{i,j},z_{e,f}),& \text{$l(i,j) = 1$}\\
z_{i,j},& \text{$l(i,j) = 0$}
\end{cases}
\end{equation}

\begin{equation}
\label{eq5}
\begin{split}
C = \sum_{z_{e,f}\in B(z_{i,j}, \ r)}u(z_{i,j},z_{e,f}), \\  \\ , u(z_{i,j},z_{e,f})=
\begin{cases}
e^{-\frac{d(z_{i,j},z_{e,f})}{h^{2}}},& \text{$e \neq i$ or $f \neq j$}\\
0,& \text{$e = i$ and $f = j$}
\end{cases}
\end{split}
\end{equation}

\begin{equation}
\label{eq6}
\begin{split}
d(z_{i,j},z_{e,f})=\left \| N(z_{i,j}) - N(z_{e,f}) \right \|^{2}_{2,a}, \\ h = (\frac{\bar{L}}{M*N})^{2} * \beta _2 + \frac{\bar{L}}{M*N} * \beta _1 + \beta _0
\end{split}
\end{equation}
where $B(z_{i,j}, \ r)$ represents a searching window of size $(2r + 1) \times (2r + 1)$ centered at $z_{i,j}$, and $u(z_{i,j},z_{e,f})$ represents the weight of pixel $z_{e,f}$ in $B(z_{i,j}, \ r)$. $N(z_{i,j})$, also called similarity window, is a square block centered at $z_{i,j}$, so is $N(z_{e,f})$. As shown in Eq. (\ref{eq6}), the similarity between $z_{i,j}$ and $z_{e,f}$ is measured by the Gaussian weighted Euclidean distance $d(z_{i,j},z_{e,f})$ between $N(z_{i,j})$ and $N(z_{e,f})$,  where $a$ is the standard deviation of the Gaussian kernel, $h$ is the smoothing parameter for NLM.

When processing the noisy pixel $z_{i,j}$, original NLM assigns the weight based on the similarity, that is to say, the weight of pixel $z_{i,j}$ itself is the largest. Different from original NLM, in our method, the noisy pixel to be processed will not participate in the process of NLM, thus the weight of the pixel $z_{i,j}$ should be set as $0$, as shown in Eq. (\ref{eq5}).

In the NLM algorithm, the higher the noise intensity is, the larger the smoothing parameter $h$ should be. But the intensity of noise is not easy to be confirmed. Considering that SAP noise can be significantly detected, we use the intensity of SAP noise to confirm $h$. As shown in Eq. (\ref{eq6}), $\bar{L}$ represents the total number of non-zero elements in the discriminant matrix $\bm{L}$, that is, the more noisy pixels are detected, the larger $h$ should be, $\beta _0, \beta _1$ and $\beta _2$ are the parameters used to fit $h$.

Due to the high computational cost of NLM algorithm, here we introduce a kind of fast implementation of NLM algorithm \cite{froment2014parameter} based on the computation of patch distances using sums of lines to accelerate our NAMF algorithm. The details of the proposed NAMF are shown in Algorithm \ref{al1}.

\begin{algorithm}[htbp]
	
	\begin{algorithmic}\\
		\label{al1}
		\caption{NAMF}
		\\/*STAGE 1*/\\
		
		\quad Compute the indicator matrix $\bm{O}$.\\
		\quad For each pixel $(i, j) \in \Lambda$ in the noisy image $\bm{Y}$ and the initially restored image $\bm{Z}$, do\\
		1) If $o_{i,j} == 0$, $l(i, j) = 0$, $z_{i, j} = y_{i,j}$, break; \\ \quad Otherwise, go to step 2).\\
		2) Initialize $w = 1$, $h = 1$, $w_{max} = 7$.\\
		3) Compute $S^{sum}_{i,j}(w)$ until $w == w_{max}$ or $S^{sum}_{i,j}(w) > 0$; \\ \quad Otherwise, $w = w + h$ and repeat step 3).\\
		4) If $S^{sum}_{i,j}(w) > 0$, $l(i, j) = 1$, $z_{i,j} =  S^{mean}_{i,j}(w)$, break; \\ \quad Otherwise, go to step 5).\\ 
		5) Compute $\rho$. If $\rho \leq  T$, $l(i, j)  =  1$, $z_{i,j} =  S^{mean}_{i,j}(w)$;\\\quad Otherwise, $l(i, j) = 0$, $z_{i,j} = y_{i,j}$.\\
		
		\\/*STAGE 2*/\\
		\quad Compute $h$. \\ \quad For each pixel $z_{i,j}$ in image $\bm{Z}$ and $\hat{z}_{i,j}$ in output image $\bm{\hat{Z}}$, do\\
		6) If $l(i, j) == 1$, $\hat{z}_{i,j} = \frac{1}{C}\sum_{z_{e,f}\in B(z_{i,j}, \ r)}z_{e,f} * u(z_{i,j},z_{e,f})$; \\ 
		\quad Otherwise, $\hat{z}_{i,j} = z_{i,j}$.
	\end{algorithmic}
\end{algorithm}

\section{EXPERIMENTAL RESULTS}

\begin{figure*}[htbp!]
	\centerline{\includegraphics[width = 1.8\columnwidth]{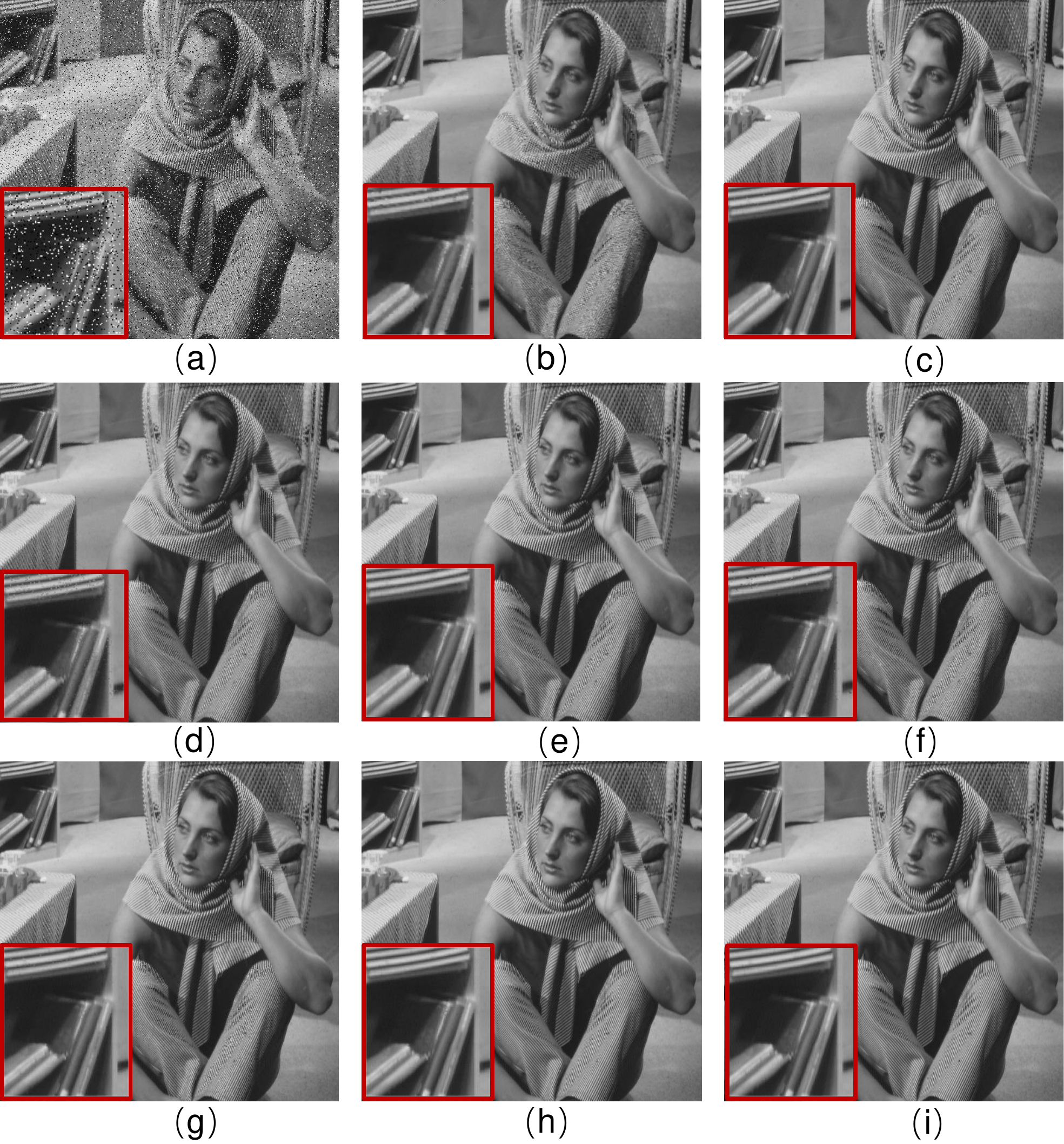}}
	\caption{Experimental results of different methods for "Barbara" with SAP noise ratio of 10\%. (a) Noisy image (15.4199 dB, 0.2586). (b) AMF (24.4698 dB, 0.7744) (c) NAFSMF (33.6841 dB, 0.9788). (d) AWMF (32.1548 dB, 0.9644). (e) \cite{Yi2016An} (34.7489 dB, 0.9820). (f) BPDF (32.6820 dB, 0.9720). (g) DAMF (33.6841 dB, 0.9788). (h) NAMF (\textbf{41.3133 dB}, \textbf{0.9932}). (i) Original image.}
	\label{fig2}
\end{figure*}

\begin{figure*}[htbp!]
	\centerline{\includegraphics[width = 1.8\columnwidth]{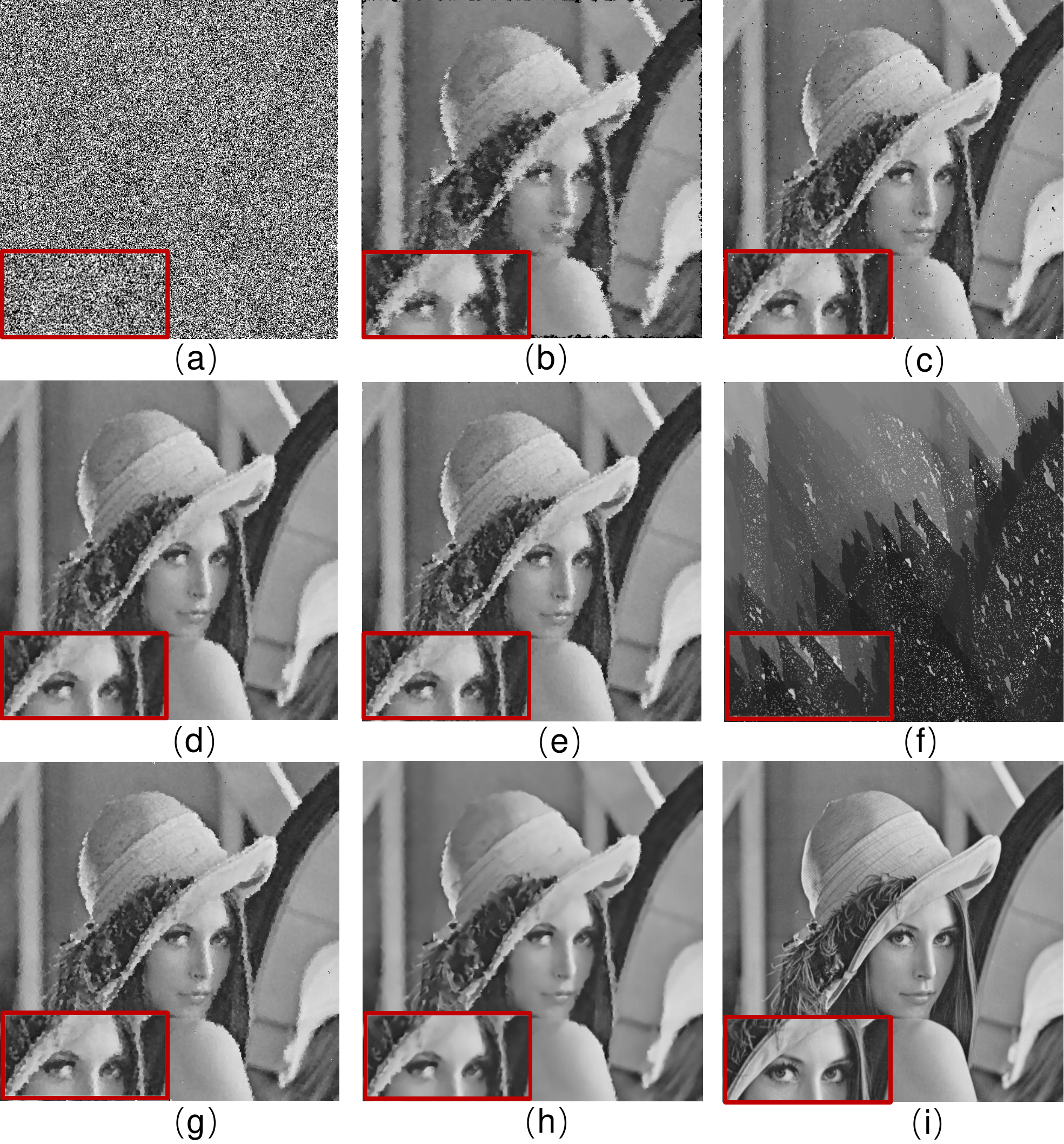}}
	\caption{Experimental results of different methods for "Lena" with SAP noise ratio of 90\%. (a) Noisy image (5.8973 dB, 0.006). (b) AMF (20.0591 dB, 0.5860) (c) NAFSMF (23.7711 dB, 0.6959). (d) AWMF (26.1224 dB, 0.7652). (e) \cite{Yi2016An} (26.1986 dB, 0.7710). (f) BPDF (10.8694 dB, 0.2775). (g) DAMF (25.9059 dB, 0.7631). (h) NAMF (\textbf{27.5748 dB}, \textbf{0.8150}). (i) Original image.}
	\label{fig2_2}
\end{figure*}

In the experiments, NAMF are compared with six state-of-the-art methods: AMF \cite{Hwang1995Adaptive}, NAFSMF \cite{Toh2010Noise}, AWMF \cite{Zhang2014A}, the method proposed in \cite{Yi2016An}, BPDF \cite{tbtkelektrik400946}, and DAMF \cite{ERKAN2018789}. Sixteen typical images (Barbara, Elaine, Goldhill, Lena, Man, Peppers, Yacht, and Zelda, Baboon, House, Boat, Cameraman, Einstein, Face, Straw, and Couple) and 40 test images in the TEST IMAGES Database \cite{testingimge} are chosen for the experiments. 

We use two typical image quality metrics, peak signal-to-noise ratio (PSNR) \cite{Hor2010Image} and structural similarity (SSIM) \cite{Zhou2004Image} to evaluate the experimental results. For an image $U$ and an image $V$ with same size of $M \times N$ , PSNR can be calculated as follows:

\begin{equation}
\begin{split}
PSNR(U, V)=10 \ * log_{10}(255^2/MSE), \\ 
MSE = \frac{1}{M*N} \sum_{i}^{M}\sum_{j}^{N}(U_{i, j} - V_{i, j})^2
\end{split}
\end{equation}
where MSE is the mean square error of two images, and $U_{i, j}$, $V_{i, j}$ are pixels of image $U$ and $V$, respectively. SSIM for image $U$ and $V$ can be defined as follows:

\begin{equation}
\begin{split}
SSIM(U, V)=\frac{(2\mu_u \mu_v + c1) * (2\sigma_{u, v} + c2) }{(\mu_u^2 + \mu_v^2 + c1)*(\sigma_u^2 + \sigma_v^2 + c2)} 
\end{split}
\end{equation}
where $\mu_u$ and $\mu_v$ are the average intensities of image $U$ and $V$, respectively. $\sigma_u$ and $\sigma_v$ are standard deviations; $\sigma_{u,v}$ is the covariance; $c1$ and $c2$ are some constants. Here $c1$ and $c2$ are set to be $(0.01*255)^2$ and $(0.03*255)^2$ as in \cite{Zhou2004Image}, respectively.

\begin{figure}[htbp!]
	\centerline{\includegraphics[width=0.825\columnwidth]{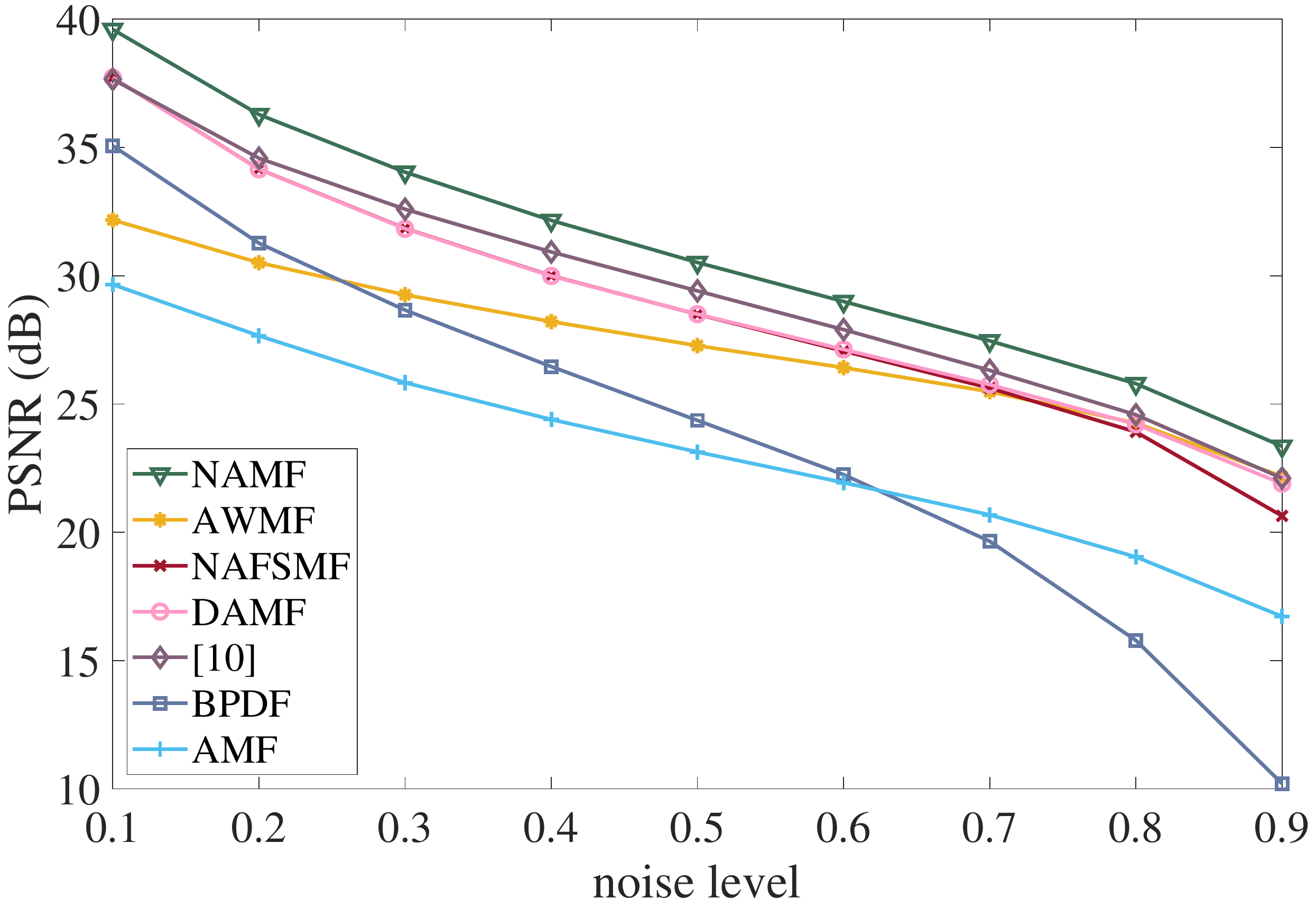}}
	\caption{Average PSNR of different methods at all SAP noise levels.}
	\label{fig4}
\end{figure}

\begin{figure}[htbp!]
	\centerline{\includegraphics[width=0.825\columnwidth]{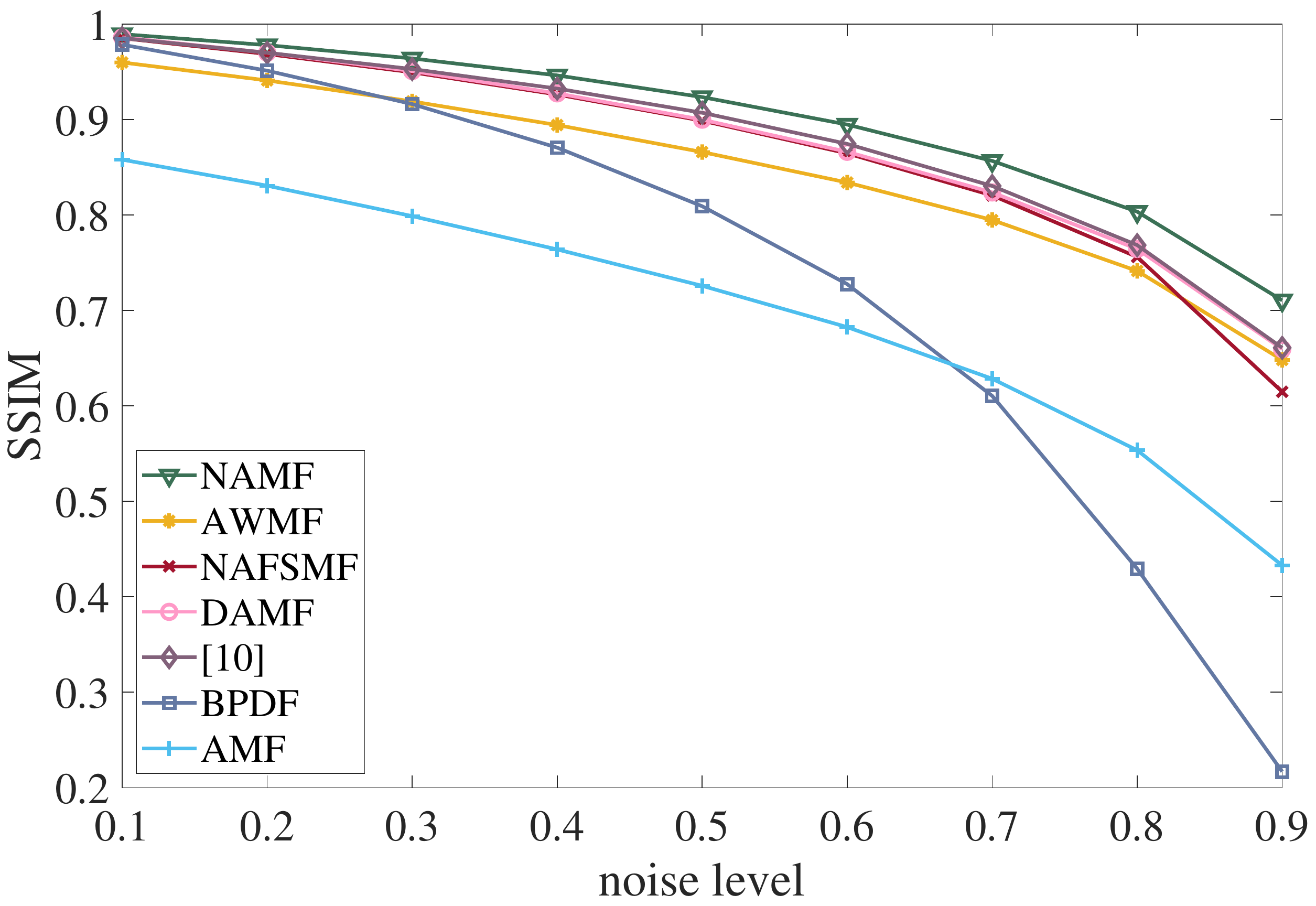}}
	\caption{Average SSIM of different methods at all SAP noise levels.}
	\label{fig5}
\end{figure}

In this paper, we set $\beta _2=2.2186$, $\beta _1 = 6.0314$ and $\beta _0 = 4.5595$ to fit $h$. The size of searching window is $5 \times 5$. And the size of similarity window is $41 \times 41$. Through test, we finally take threshold $T = 0.8$ for our method. Other methods keep the default parameters. The experiments are performed on a personal computer with Intel Core i7 2.2 GHz processor and 16 GB RAM.

Fig. \ref{fig2} shows the restored results of all methods for "Barbara" with SAP noise ratio of 10\%. By observing the enlarged area in Fig. \ref{fig2}, we can see that AMF, AWMF, and BPDF can't restore the details of the original image very well. And the result of our proposed NAMF is nearly the same as the original image.

Fig. \ref{fig2_2} shows the restored results for "Lena" with SAP noise ratio of 90\%. In the enlarged area in Fig. \ref{fig2_2}, it can be found that the performances of AMF, NAFSMF, and BPDF are very poor. And the restored images by other methods are also very blurred, while result of our method looks more natural and smooth.

The curves of average PSNR and SSIM are shown in Fig. \ref{fig4} and Fig. \ref{fig5} respectively. Fig. \ref{fig4} illustrates that NAMF obtains the highest PSNR under both low and high noise intensity, and PSNR of our method is much higher than results of other methods. Fig. \ref{fig5} shows that the SSIM curves of most methods are basically the same under low SAP noise intensity. However, with the increasing of noise level, the superiority of NAMF is getting more obviously. After the noise ratio exceeding 30\%, the SSIM obtained by NAMF is significantly higher than other methods.

\begin{figure}[htbp!]
	\centerline{\includegraphics[width=0.825\columnwidth]{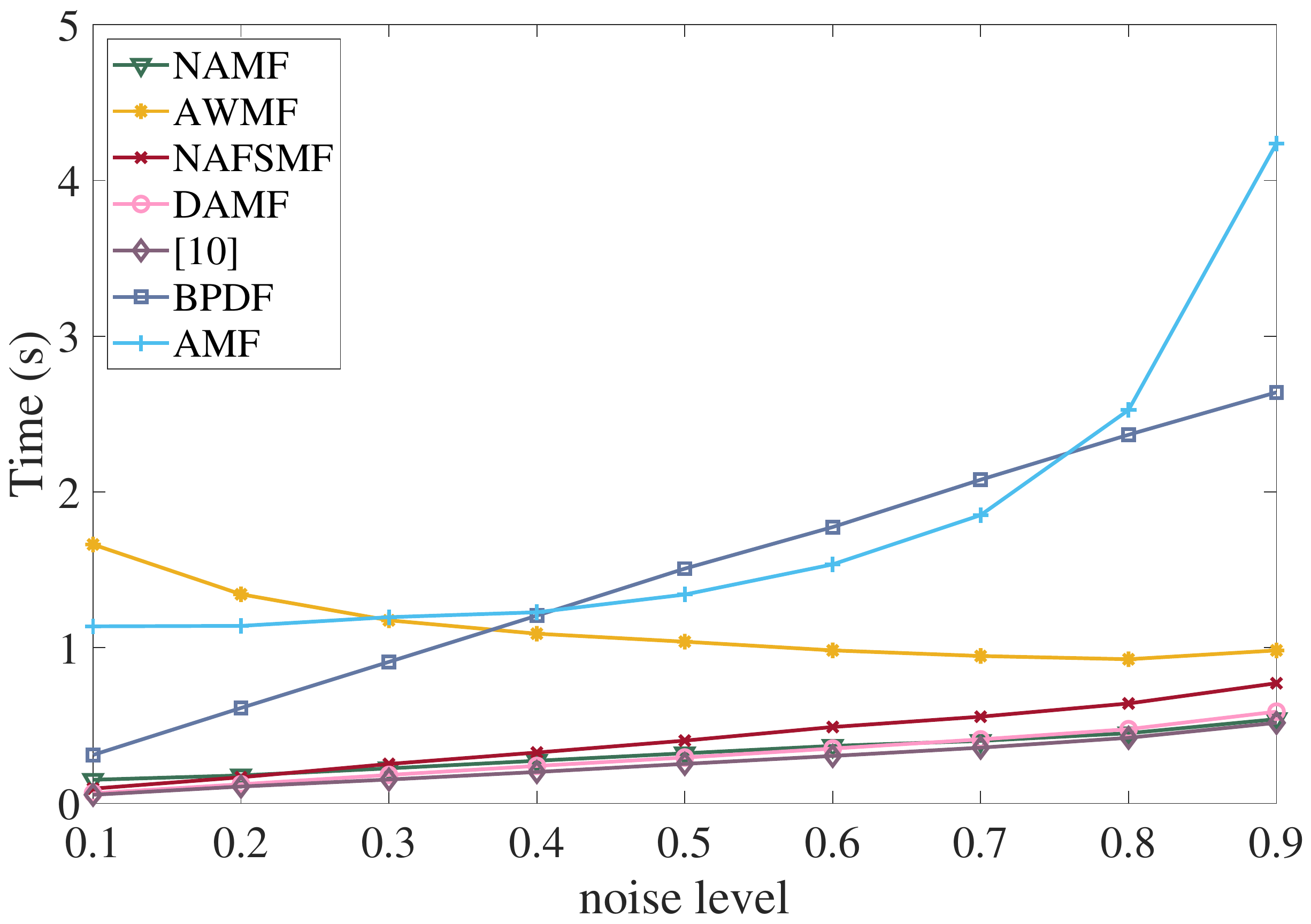}}
	\caption{Average running time of different methods at all SAP noise levels.}
	\label{fig6}
\end{figure}

Fig. \ref{fig6} illustrates the average running time of different methods at all noise levels. With the increasing of noise intensity, the running time of all methods except AWMF increases. Clearly, the average running time of the method proposed in \cite{Yi2016An} is the shortest. Although the rank of NAMF is in the middle, its processing speed is superior to NAFSMF and similar with DAMF under high noise intensity.

\section{Conclusion}

In this paper, a method called NAMF for SAP noise denoising is proposed, which adopts a SAP noise based non-local mean method. NAMF can get much higher restoring quality than state-of-the-art methods at all SAP noise levels. The processing time of NAMF is comparable to most state-of-the-art methods. The experimental results show that NAMF can get much better PSNR and SSIM at all SAP noise levels. Moreover, NAMF can preserve more details even at noise level as high as 90\%.

\bibliographystyle{IEEEbib}
\bibliography{mybib}

\end{document}